\documentclass[review]{elsarticle}

\usepackage[english]{babel}
\usepackage[T1]{fontenc}
\usepackage{etoolbox}
\makeatletter
\patchcmd{\ps@pprintTitle}{\footnotesize\itshape
      Preprint submitted to \ifx\@journal\@empty Elsevier
      \else\@journal\fi\hfill\today}{\scriptsize{Preprint submitted to Solar Energy \hfill \today}}{}{}

\def\ps@pprintTitle{%
   \let\@oddhead\@empty
   \let\@evenhead\@empty
   \let\@oddfoot\@empty
   \let\@evenfoot\@oddfoot
}

\makeatother
\usepackage{comment}
\usepackage{subfig}
\usepackage{hyperref}
\usepackage{amsmath}

\DeclareMathOperator*{\argmin}{arg\,min}

\usepackage{amsmath,amssymb,amsfonts}
\usepackage{algorithmic}
\usepackage[]{graphicx}
\graphicspath{{Figures/}}
\usepackage{lipsum}
\usepackage{balance}
\usepackage{textcomp}
\usepackage{xcolor}
\usepackage{multirow}
\usepackage{color}
\usepackage{lscape}
\usepackage{longtable,booktabs}
\usepackage[switch]{lineno}
\usepackage{seqsplit}
\begin{document}

\begin{frontmatter}

\title{DMCNet: Diversified Model Combination Network for Understanding Engagement from Video Screengrabs} 

\author[add1]{Sarthak~Batra\corref{contrib}}
\ead{sarthak.batra@adaptcentre.ie}
\author[add2]{Hewei~Wang\corref{contrib}}
\ead{hewei.wang@ucdconnect.ie}
\author[add3]{Avishek~Nag}
\ead{avishek.nag@ucd.ie}
\author[add4]{Philippe~Brodeur}
\author[add5]{Marianne~Checkley}
\author[add6]{Annette~Klinkert}
\author[add1,add7]{Soumyabrata~Dev\corref{mycorrespondingauthor}}
\cortext[contrib]{Authors contributed equally to this research.}
\cortext[mycorrespondingauthor]{Corresponding author. Tel.: + 353 1896 1797.}
\ead{soumyabrata.dev@ucd.ie}

\address[add1]{ADAPT SFI Research Centre, Dublin, Ireland}
\address[add2]{Beijing University of Technology, Beijing, China}
\address[add3]{School of Electrical and Electronic Engineering, University College Dublin, Ireland}
\address[add4]{Overcast, Dublin, Ireland}
\address[add5]{Camara Education, Dublin, Ireland}
\address[add6]{European Science Engagement Association, Vienna, Austria}
\address[add7]{School of Computer Science, University College Dublin, Ireland}

\begin{abstract}
Engagement is an essential indicator of the Quality-of-Learning Experience (QoLE) and plays a major role in developing intelligent educational interfaces. The number of people learning through Massively Open Online Courses (MOOCs) and other online resources has been increasing rapidly because they provide us with the flexibility to learn from anywhere at any time. This provides a good learning experience for the students. However, such learning interface requires the ability to recognize the level of engagement of the students for a holistic learning experience. This is useful for both students and educators alike. However, understanding engagement is a challenging task, because of its subjectivity and ability to collect data. In this paper, we propose a variety of models that have been trained on an open-source dataset of video screengrabs. Our non-deep learning models are based on the combination of popular algorithms such as Histogram of Oriented Gradient (HOG), Support Vector Machine (SVM), Scale Invariant Feature Transform (SIFT) and Speeded Up Robust Features (SURF). The deep learning methods include Densely Connected Convolutional Networks (DenseNet-121), Residual Network (ResNet-18) and MobileNetV1. We show the performance of each models using a variety of metrics such as the Gini Index, Adjusted F-Measure (AGF), and Area Under receiver operating characteristic Curve (AUC). We use various dimensionality reduction techniques such as Principal Component Analysis (PCA) and t-Distributed Stochastic Neighbor Embedding (t-SNE) to understand the distribution of data in the feature sub-space. Our work will thereby assist the educators and students in obtaining a fruitful and efficient online learning experience.
\end{abstract}

\begin{keyword}
Engagement \sep Facial Expression Recognition \sep Deep Learning \sep Convolutional Neural Network \sep Histogram of Oriented Gradient \sep Support Vector Machine
\end{keyword}

\end{frontmatter}

\section{Introduction}
\label{sec:1}

\begin{figure}[htbp]
\includegraphics[width=2cm, height=2cm]{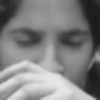}
\includegraphics[width=2cm, height=2cm]{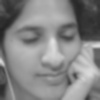}
\includegraphics[width=2cm, height=2cm]{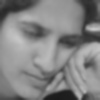}
\centering

\includegraphics[width=2cm, height=2cm]{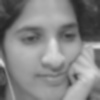}
\includegraphics[width=2cm, height=2cm]{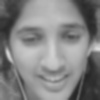}
\includegraphics[width=2cm, height=2cm]{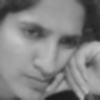}
\centering

\includegraphics[width=2cm, height=2cm]{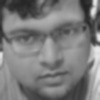}
\includegraphics[width=2cm, height=2cm]{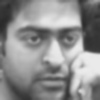}
\includegraphics[width=2cm, height=2cm]{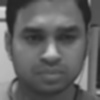}
\centering
\caption{Disengaged (top), partially engaged (middle) and engaged (bottom) samples of the dataset.}
\label{fig:dataset-introduction}
\end{figure}

Engagement is an important part of human-technology interactions and is defined
differently for a variety of applications such as search engines, online gaming
platforms, and mobile health applications. Most definitions describe engagement
as attention and emotional involvement in a task. This paper deals with
engagement during learning via technology. Investigating engagement is vital
for designing intelligent educational interfaces in different learning
platforms including educational games, MOOCs, and Intelligent Tutoring Systems
(ITSs). For instance, if students feel frustrated and become disengaged (see
disengaged samples in Fig.~\ref{fig:dataset-introduction}), the system should intervene in order to bring
them back to the learning process. However, if students are engaged and
enjoying their tasks (see engaged samples in  Fig.~\ref{fig:dataset-introduction}), they should not be
interrupted even if they are making some mistakes. In order for the learning
system to adapt the learning setting and provide proper responses to students,
we first need to automatically measure engagement. In order to perform this, we train a few models
\textit{viz.} a CNN, a SVM, and deep learning methods~\cite{islam2020foreign} to make
correct predictions about the involvement of the student. In this
paper,\footnote{In the spirit of reproducible research, all the codes related to
this manuscript is available via:
\url{https://github.com/WangHewei16/Video-Engagement-Analysis}.} we describe the various problems that we faced during our project and how we overcame them. We also show different techniques of extracting features from images and how we trained our machine-learning-based models.

Engagement analysis has become more important during these difficult times of COVID-19 pandemic. As in most of the countries the college lectures are being taken online so it becomes important for the teachers to get feedback about engagement of students during those lectures. This will greatly assist the teachers to figure out why and where are students getting disengaged. This will also help teachers to give more attention to students that are disengaged most of the times. Current research on this technology of engagement analysis exhibit the following two major gaps:
\begin{itemize}
    \item Lack of diversification of models and lack of efficient algorithms to detect engagement, which makes it difficult for users to find the most suitable model for the chosen platform.
    \item There exists few techniques and objective measures to evaluate performance of the machine-learning models, making it difficult to benchmark results.  
\end{itemize}

To solve the aforementioned issues, we propose various non-deep learning based models and deep learning based models. The non-deep-learning models are based on the combination of popular algorithms \textit{viz.} HOG, SVM, SIFT and SURF. The deep learning methods include DenseNet-121, ResNet-18 and MobileNetV1. In terms of the experiment, we have used several metrics \textit{viz.} Gini Index, Adjusted F-score (AGF), Area Under Curve (AUC) in order to show each model's performance. We also use dimensionality reduction techniques to illustrate the distribution of the experimental data.

The main contributions of this paper include:
\begin{itemize}
    \item We propose various models that have been trained to identify user engagement on a given dataset and explain in detail how each model was prepared and trained;
    \item Metrics such as Gini Index, AGF, AUC are used to demonstrate the performance of each model while dimension reduction techniques such as principal component analysis (PCA), t-Distributed Stochastic Neighbor Embedding (t-SNE) are used to visualise the distribution of data;
    \item All the codes related to this study are open-source to facilitate researchers' repeatable experiments.
\end{itemize}

This paper is organized as follows.  Section~\ref{sec:3} analyzes and discusses the relevant research of this field.  Section~\ref{sec:Engagement Analysis Datasets} introduces the  WACV dataset which we used in our research. Section~\ref{sec:4} describes the several models that are proposed.  Section~\ref{sec:Results} conducts experiments and discusses the results. Section~\ref{sec:5} concludes this paper with future work.

\section{Related Work}

\label{sec:3}
\subsection{Research background}

In the past, research focused on using machine learning methods in the
development of the personalization of curriculum, adaptive evaluation and
recommendation systems based on learner preferences and browsing behavior,
such as
in~\cite{rigou2004integrating,huang2007constructing,baylari2009design,dolog2004personalization}.
Whilst in the recent times, there has been an exponential growth of using
MOOCs~\cite{qi2021evaluating}. They have laid the path for modern education
where students can learn whatever they want and from wherever they want to
learn. Obtaining students' engagement in an online class is a core issue of
this study, as it is important to keep students focused in the era of online
education. After obtaining the engagement of students, educators can better
supervise students' learning efficiency, and teachers can also improve their
teaching methods according to the concentration and engagement data of
students. Our research has had a positive impact on this modernization
education issue.

\subsection{Development of facial expression recognition technology}

Facial expression recognition technology plays an essential role in engagement
recognition. In~\cite{kolekar2010learning}, the authors achieved facial
expression recognition technique based on machine learning algorithms called
Artificial Neural Networks (ANNs) and Web Usage Mining (WUM) to create an
adaptive e-learning environment. One of the important aspects that are
currently missing from these online education platforms is that teachers do not
get feedback on how attentive the student was based on human's reactions,
expressions, and position. To tackle this problem researchers adopted various
machine learning and computer vision techniques. The authors have proposed a
part-based hierarchical bidirectional recurrent neural network (PHRNN) to
analyze the facial expression information of temporal sequences. PHRNN models
facial morphological variations and dynamical evolution of expressions, which
is effective to extract ``temporal features'' based on facial landmarks. Use of
Computer Expression Recognition Toolbox (CERT)
in~\cite{grafsgaard2013automatically} to track fine-grained facial movements
consisting of eyebrow-raising (inner and outer), brow lowering, eyelid
tightening, and mouth dimpling within a naturalistic video corpus of tutorial
dialogue (N $=$ 65). Within the dataset, upper face movements were found to be
predictive of engagement, frustration, and learning, while mouth dimpling was a
positive predictor of learning and self-reported performance. These results
highlight how both intensity and frequency of facial expressions predict
tutoring outcomes, use of HOG and SVM to detect humans has been implemented
in~\cite{dalal2005histograms}. In~\cite{hernandez2013measuring}, Hernandez
\textit{et al.}  modeled the problem of determining the engagement of a TV viewer as a
binary classification problem, using multiple geometric features extracted from
the face and head and used SVMs for the classification.
In~\cite{zhang2015learning}, the authors have used deep-learning-based
techniques to classify facial images into different social relation traits,
facial expressions are used in~\cite{zhang2017facial}.
In~\cite{bosch2016using}, the authors used computer vision and machine-learning
techniques to detect students' effects from facial expressions (primary
channel) and gross body movements (secondary channel) during interactions with
an educational physics game. We use one of the publicly available datasets that
are provided by~\cite{kamath2016crowdsourced}.
In~\cite{pramerdorfer2016facial}, the authors achieve facial expression
detection using CNN based approach. In~\cite{chen2017understanding}, the
authors design a user engagement prediction framework and specific video
quality metrics to help the providers of content, predict the time, viewers
tend to remain in video sessions. In~\cite{knyazev2018leveraging}, the authors
introduce a way to detect the emotion of the people in videos and make his
accuracy greater than 60\% with high reliability through the industry-level
face recognition networks.  In~\cite{qiao2019automated}, the authors introduces
a preliminary model, which can help students make course choices faster and
provide better-personalized services. However, how to provide a better
personalized and adaptive e-learning environment is still a challenge that has
attracted many researchers. In~\cite{megahed2020modeling}, a coupled
integration between CNN and a fuzzy system is adopted. The CNN is used to
detect a learner's facial expressions and a fuzzy system is used to determine
the next learning level based on the extracted facial expression states from
the CNN and several response factors by the learner.

\subsection{Development of engagement recognition technology and research motivation}

In~\cite{monkaresi2016automated}, features like heart rate, Animation Units
(from Microsoft Kinect Face Tracker), and local binary patterns in three
orthogonal planes (LBP-TOP) were used in supervised learning for the detection
of concurrent and retrospective self-reported engagement. Theoretical analysis
of engagement done in~\cite{o2016theoretical,bosch2016detecting}, talks about
intrinsic motivation, flow theory, focus attention, positive psychology and
social enterprise. These studies show some factors and theories that influence
the engagement of humans. Use of facial features done in~\cite{alyuz2016semi}
are used to automatically detect student engagement and the result demonstrated
initial success in this field. An important factor in solving challenging
problems like determining engagement requires a well-annotated dataset.
In~\cite{aslan2017human}, the authors have discussed for affective state
annotation, how does the socio-cultural background of human expert labelers,
compared to the subjects, impact the degree of consensus and distribution of
affective states obtained and Secondly, how do differences in labeler
background impact the performance of detection models that are trained using
these labels. In~\cite{nezami2019automatic}, the authors present a deep
learning model to improve engagement recognition from images that overcome the
data sparsity challenge by pre-training on readily available basic facial
expression data, before training on specialized engagement data. In the first
step, a facial expression recognition model is trained to provide a rich face
representation using deep learning. In the second step, they used the model's
weights to initialize the deep-learning-based model to recognize engagement.
The most common facial expressions used are (happiness, sadness, anger,
disgust, fear, surprise). There is some interesting research such
as~\cite{lin2020face}, where the authors introduced a recognition model with
high accuracy which uses t-SNE to improve the performance of the model. The
emotion of people can also be analyzed. Nezami \textit{et al.}  have performed some
relevant researches in~\cite{nezami2018deep,nezami2018automatic}, the authors
use some deep learning methods based on CNNs model to recognize facial
expression and engagement. Current engagement recognition research, on the
other hand, reveals two major flaws. The first gap is a lack of model diversity
as well as efficient model creation procedures. Users will have a tough time
finding the best model for the platform as a result of this. The second flaw is
that there are just a few types of assessment criteria for analyzing model
performance, making it difficult to come up with compelling findings. We are
inspired to propose alternative models, including non-deep learning and deep
learning models, to cover these gaps. Non-deep learning approaches rely on a
mix of several computer vision features. The deep learning methods include
DenseNet-121, ResNet-18 and MobileNetV1. The reason for using these algorithms
on the engagement detection datasets is because they are well-known and widely
used techniques in the face recognition field. As a result, we are interested
in systematically merging these technologies so that we can clearly examine the
influence and performance of each model, and so choose the optimal model
through several tests and experiments. In terms of the experiment, we employ
numerous measures to illustrate the performance of each model, including the
Gini Index, AGF, and AUC. We also employ dimensionality reduction techniques
such as PCA and t-SNE to depict the distribution of experiment data in a more
thorough and varied manner.  

In terms of the managerial insights, our research team, online education platform and users need to reach an agreement on privacy and profitability, and then reasonably apply the model on the platform to test students' engagement.

\section{Engagement Analysis Datasets}
\label{sec:Engagement Analysis Datasets}
This section introduces the related datasets in this field and describe WACV dataset which we used in our research. The public datasets available are {\bf DAiSEE dataset} which consists of four classes (engaged, frustration, boredom, confusion) where images are generated from 9068 videos and 112 subjects (80 male and 32 female), {\bf HBCU dataset} also consists of four classes (not engaged, nominally-engaged, engaged, very-engaged) where images are generated from 120 videos and 34 subjects (9 male and 25 female). {\bf In-the-wild} has classes (disengaged, barely-engaged, normally-engaged, highly-engaged) where images are generated from 195 videos and 78 subjects (53 male and 25 female). {\bf SDMATH dataset} has been generated from 20 videos and 20 subjects (10 males and 210 females).

Now we discuss how we analyzed the engagement of students. We used the {\bf WACV dataset} which is a public dataset for our work. The dataset has three different classes that are disengaged, partially engaged, and engaged. It contains a total of 4424 3-channel images of varying sizes. We reshape all the images to a same shape of (100, 100, 3). The dataset is not a balanced dataset as it has 412 images belonging to the class `disengaged', 2247 images belonging to the class `partially engaged', and 1765 images belonging to the class `engaged'. We firstly randomly selected 412 images from class 2 (partially engaged) and class 3 (engaged) so that each class has 412 images each. We split this data into training (80\%) and testing (20\%). We used different models to fit this balanced data and finally chose the one which provided the best test accuracy. Figure~\ref{fig:dataset-introduction} shows the three classes of WACV dataset. 
The link to the dataset that we have used is  \url{https://github.com/e-drishti/wacv2016}.

In~\cite{whitehill2014faces}, the authors used HBCU dataset for the automatic
detection of learners' engagement from facial expressions. As mentioned
earlier, this research study compared three machine learning techniques ---
Boost (BF), SVM (Gabor), and MLR (CERT) (Whitehill \textit{et al.}  2014). Four fold
subject-independent cross-validation with the 2AFC was done to measure the
accuracy for engagement detection. The average accuracies achieved by the MLR
(CERT), Boost (BF), and the SVM (Gabor) were 0.714, 0.728 and 0.729,
respectively. In~\cite{sathayanarayana2014towards}, the authors used the SDMATH
dataset to detect the deictic tip for the hand gesture recognition. This method
employed graph based visual saliency (GBVS) and SVM to detect the deictic tip,
and achieves the TPR of 85\%. In~\cite{gupta2016daisee}, the authors used
DAiSEE dataset for engagement detection through using three different models of
Convolution Neural Networks (CNNs) --- InceptionNet, C3D, and Long-Term
Recurrent Convolutional Network (LRCN). The models were applied to detect
boredom, engagement, confusion, and frustration, where InceptionNet achieved
the accuracies of these engagement levels 36.5\%, 47.1\%, 70.3\%, and 78.3\%,
respectively. The C3D and LRCN achieved the accuracies of these engagement
levels of 45.2\%, 56.1\%, 66.3\%, 79.1\%, and 53.7\%, 61.3\%, 72.3\%, 73.5\%,
respectively. In~\cite{kaur2018prediction}, the authors used the
``in-the-wild'' dataset for their performance evaluation. This method employed
three-fold cross validation with multiple kernel learning (MKL) SVM and the
average accuracy and the maximum accuracy obtained 43.98\% and 50.77\%,
respectively.

\section{DMCNet for Engagement Detection from Videos}
\label{sec:4}
\subsection{CNN based approach}

\begin{figure}[htbp]
   \centering
   \includegraphics[width=1\textwidth]{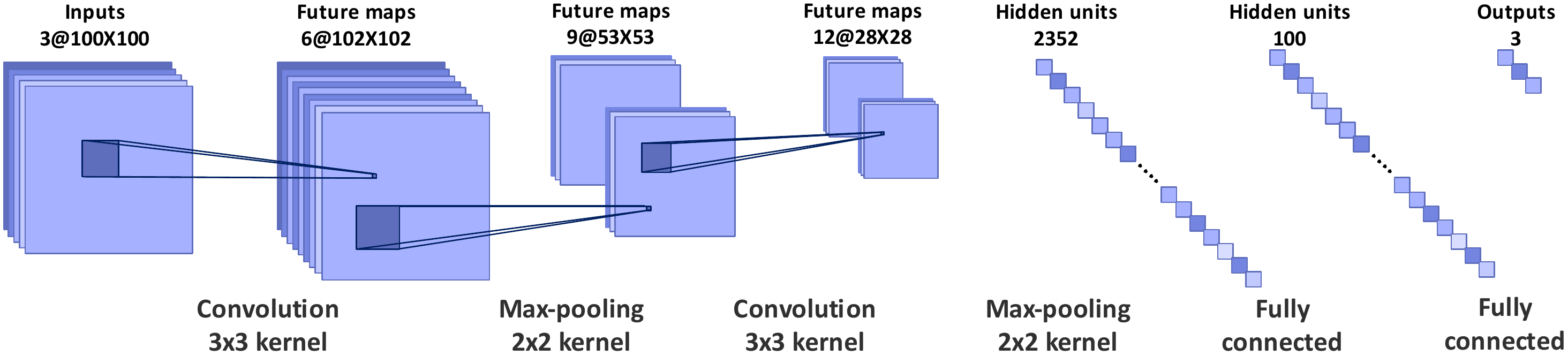}
   \caption{Architecture overview of the CNN based model}
   \label{fig:Architecture}
\end{figure}

As the training data is really small it does not make sense to use a very deep neural network as it can lead to overfitting so we used a CNN model that has three convolutional layers and each convolutional layer has 3-6-9 kernels respectively with non-linearity as relu and each convolutional layer is followed by a MaxPooling with stride 2 and finally there is a fully connected layer connected to a softmax layer. We chose MSE Loss function as our criterion and used Adam optimizer with learning rate $10^{-4}$. We did 20 experiments, in each experiment we generated our balanced data using random sampling and trained our model for 10 epochs. After training, we tested it on the test data and stored the accuracy in a list and after all experiments were completed we calculated the average accuracy.  Fig.~\ref{fig:Architecture} shows the flowchart of the CNN based model.

\subsection{{{HOG}}}

Local object appearance and shape can often be characterized rather well by the distribution of local intensity gradients or edge detection. HOG features are calculated by taking orientation histograms of edge intensity in local region.

In this paper, we extract HOG features from 8x8 local regions. At first, edge gradients and orientations are calculated at each pixel in this local region. Sobel filters are used to obtain the edge gradients and orientations.
The gradient magnitude $mag(x, y)$ and orientation $\phi(x, y)$ are calculated using the x- and $y$-directional gradients $dx(x, y)$ and $dy(x, y)$ computed by Sobel filter as:

\begin{equation}
 mag(x,y) = \sqrt{dx(x, y)^2 + dy(x, y)^2}
\end{equation}

\begin{equation}
  \phi(x,y) =
    \begin{cases}
        tan^{-1}(\frac{dy(x,y)}{dx(x,y)}) - \pi & \text{if dx(x,y) < 0 and dy(x,y) < 0}\\
        tan^{-1}(\frac{dy(x,y)}{dx(x,y)}) + \pi & \text{if dx(x,y) < 0 and dy(x,y) > 0}\\
        tan^{-1}(\frac{dy(x,y)}{dx(x,y)}) & \text{otherwise}
    \end{cases}       
\end{equation}

\begin{figure}[htbp]
\includegraphics[width=2cm, height=4cm]{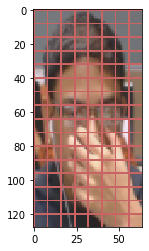}
\includegraphics[width=4cm, height=4cm]{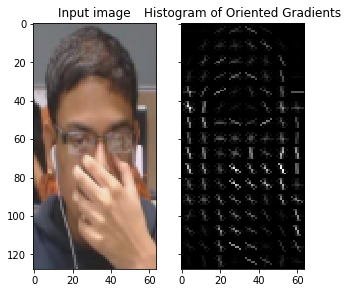}
\includegraphics[width=4cm, height=4cm]{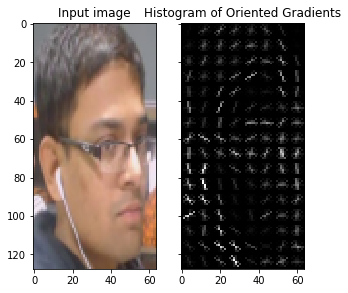}
\centering
\caption{HOG features are extracted from all grid cells of size 8x8 (left), features extracted for the image that belongs to partially engaged class (middle) and features extracted for the image that belongs to engaged class (right).}
\label{fig:HOG-features}
\end{figure}

This local region is divided into small spatial area called ``cell''. The size of the cell is 8×8 pixels. Histograms of edge gradients with 9 orientations are calculated from each of the local cells. Then the total number of HOG features becomes 576 = 9 × (8 × 8) and they constitute a HOG feature vector. To avoid sudden changes in the descriptor with small changes in the position of the window, and to give less emphasis to gradients that are far from the center of the descriptor, a Gaussian weighting function with $\sigma$ equal to one half the width of the descriptor window is used to assign a weight to the magnitude of each pixel. A HOG feature vector represents local shape of an object, having edge information at plural cells. In flatter regions like a ground or a wall of a building, the histogram of the oriented gradients has flatter distribution. On the other hand, in the border between an object and background, one of the elements in the histogram has a large value and it indicates the direction of the edge. Although the images are normalized to position and scale, the positions of important features will not be registered with same grid positions. It is known that HOG features are robust to the local geometric and photometric transformations. If the translations or rotations of the object are much smaller than the local spatial bin size, their effect is small.  Fig.~\ref{fig:HOG-features} shows the HOG features extracted from all grid cells of size {8x8}, partially engaged class and engaged class.

\subsection{HOG+SVM}

\begin{figure}[htbp]
   \centering
   \includegraphics[width=1\textwidth]{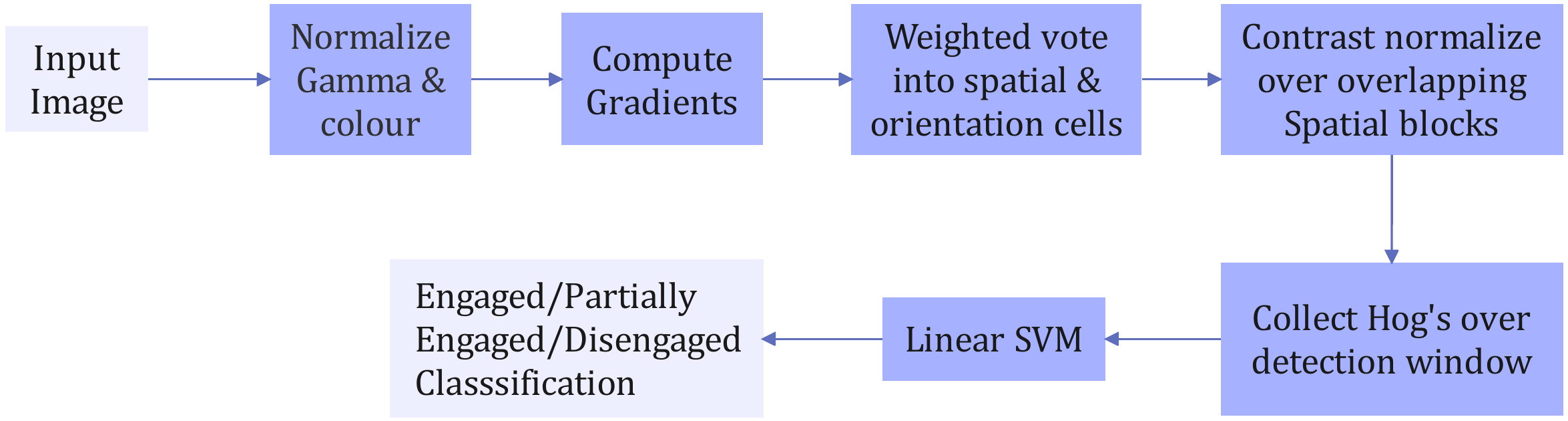}
   \caption{Architecture overview of the HOG+SVM model}
   \label{fig:Architecture-HOG-SVM-model}
\end{figure}

As the name suggests we use HOG Algorithm to extract a feature vector for each training image. Once we have done that we have created a representation of each image into a vector with 1236 elements. Now to perform the classification task we train a SVM model on these feature vectors. After training we use the test data to calculate the performance metrics.  Fig.~\ref{fig:Architecture-HOG-SVM-model} shows the flowchart of the HOG$+$SVM model.

\subsection{HOG+CNN}

\begin{figure}[htbp]
   \centering
   \includegraphics[width=1\textwidth]{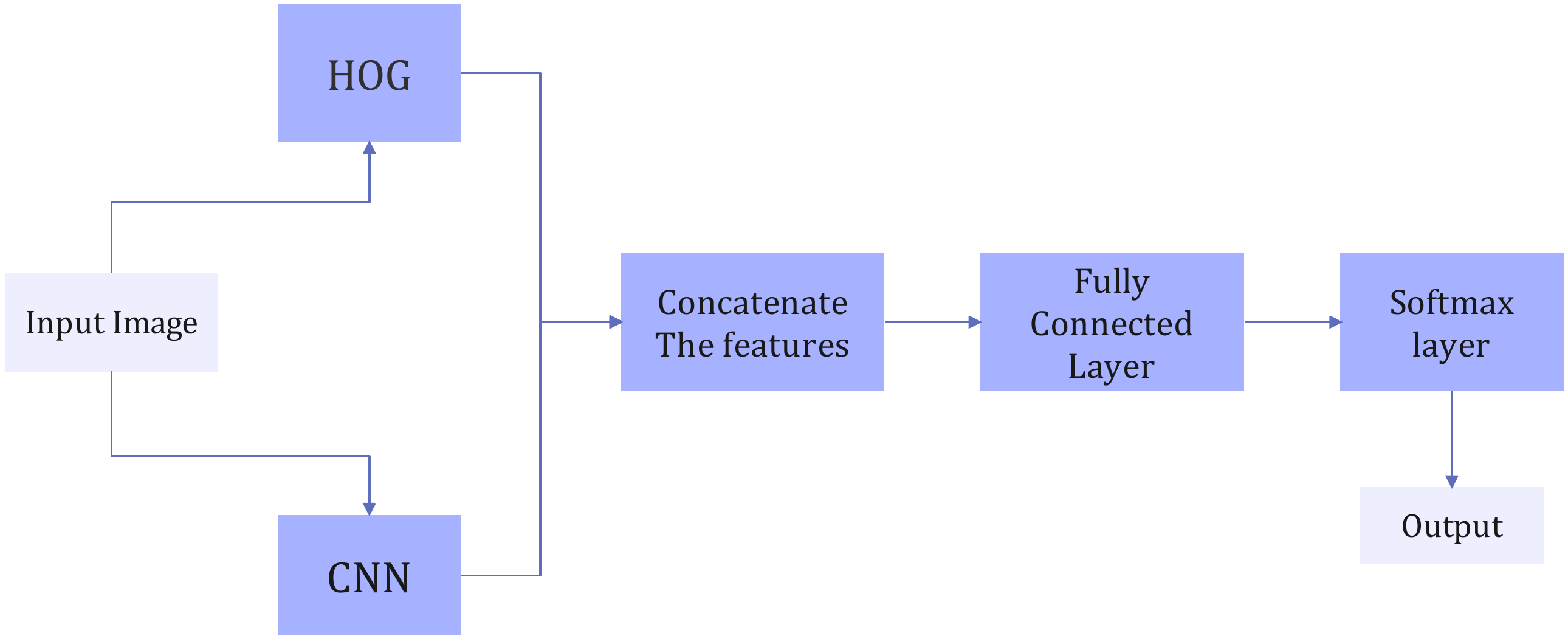}
   \caption{Architecture overview of the HOG+CNN model}
   \label{fig:Architecture-HOG-CNN}
\end{figure}

In this technique we firstly extract features vector of size 1236 using HOG and we pass the same image through the CNN and extract the feature vector from the last max pooling layer and then we concatenate the features extracted from the CNN and HOG which then goes through a fully connected layer followed by \textit{softmax} output.  Fig.~\ref{fig:Architecture-HOG-CNN} shows the flowchart of the HOG$+$CNN model.

\subsection{HOG+SIFT+SVM}

\begin{figure}[htbp]
\begin{center}
\includegraphics[height=4cm]{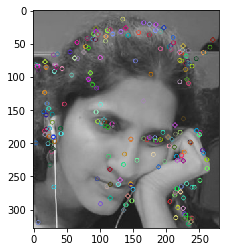}
\includegraphics[height=4cm]{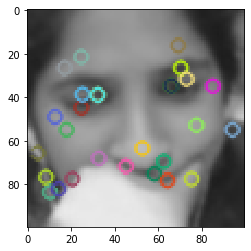}
\end{center}
\caption{Key-points detected by SIFT on a couple of training images}
\label{fig:SIFT}
\end{figure}

SIFT is a technique to detect keypoints from an image. Each keypoint has a location, scale, and  orientation. SIFT detects many keypoints but we keep the top 10 keypoints. Each keypoint is described with a descriptor. Next is to compute the descriptor for the local image region about each keypoint that is highly distinctive and invariant as possible to variations such as changes in viewpoint and illumination. To do this, a {16x16} window around the keypoint is taken. It is divided into 16 sub-blocks of {4x4} size. For each sub-block, a 8-bin orientation histogram is created. So {4x4} descriptors over {16x16} sample array were used in practice. 4x4x8 directions give 128 bin values. It is represented as a feature vector to form keypoint descriptor. We extract features using the HOG algorithm and keypoints descriptors are detected by SIFT algorithm and then we concatenate the features extracted using HOG and keypoints detected by SIFT and then fit a SVM classifier.  Fig.~\ref{fig:SIFT} shows the key-points detected by SIFT on a couple of training images.

\subsection{SURF+SVM}
\begin{figure}[htbp]
\begin{center}
\includegraphics[height=4cm]{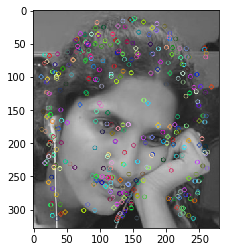}
\includegraphics[height=4cm]{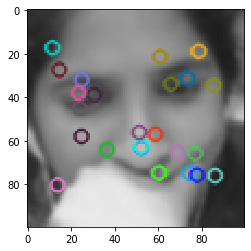}
\end{center}
\caption{Key-points detected by SURF on a couple of training images}
\label{fig:SURF}
\end{figure}

SURF is a fast and robust algorithm for local, similarity invariant representation and comparison of images. To detect interest points, SURF uses an integer approximation of the determinant of Hessian blob detector, which can be computed with 3 integer operations using a precomputed integral image. Its feature descriptor is based on the sum of the Haar wavelet response around the point of interest. These can also be computed with the aid of the integral image. Given a training image $x_i$ firstly the key-points are detected and then descriptors for each of the key-points are extracted using SURF, so for all the images present in our training data we extract the descriptors and then SVM classifier is fit on the extracted feature descriptors.  Fig.~\ref{fig:SURF} shows the key-points detected by SURF on a couple of training images.

\subsection{Deep Learning Method} 
The Deep learning method is widely used in image recognition tasks, in our project we test three kinds of image recognition models on our dataset which are DenseNet-121, ResNet-18, and MobileNetV1. In each experiment, we use 100 * 100 grayscale images as input and use the same configuration. We use Adam as our optimizer with a learning rate of 1e-5. We set the batch size to 16 and training for 10 epochs. The training and validation datasets are spit with the ratio of 8:2. After the training procedure, our model is tested on the validation set.

\subsubsection{D{enseNet}-121} 
\begin{figure}[htbp]
\centering
\begin{center}
\includegraphics[height=5.5cm]{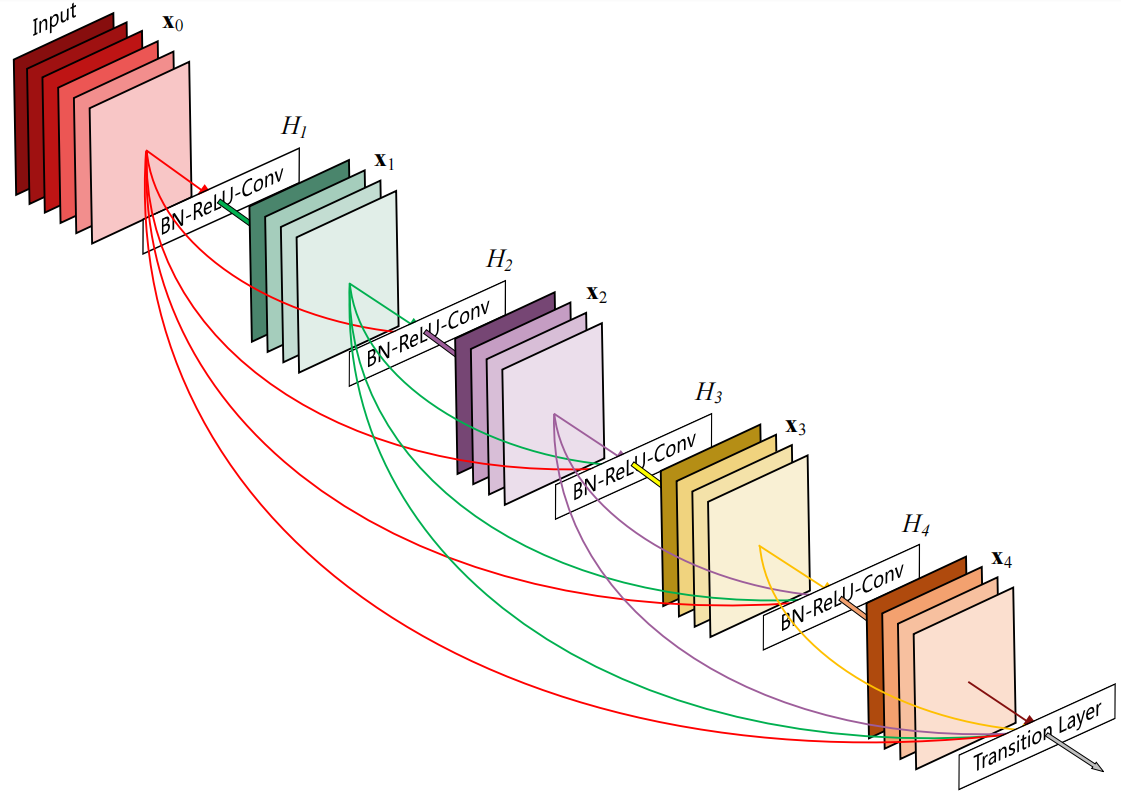}
\end{center}
\caption{A 5-layers Dense Block \cite{2016Densely}}
\label{fig:Dense-Block}
\end{figure}

DenseNet-121 have 121 convolutional layers. The DenseNet model can be seen as
the variation of the ResNet model because of the dense residual connection
between each convolutional layer in a single dense block. The
paper~\cite{2016Densely} illustrates that the dense connection between multiple
layers can preserve feed-forward feature and each layer obtains additional
inputs from all preceding layers and passes on its feature-maps to all
subsequent layers and it has fewer parameters than the model proposed by
paper~\cite{2016Deep}.  Fig.~\ref{fig:Dense-Block} shows the architecture of a 5-layers
dense block.

\subsubsection{{ResNet}-18} 
\begin{figure}[htbp]
\begin{center}
\includegraphics[height=3.5cm]{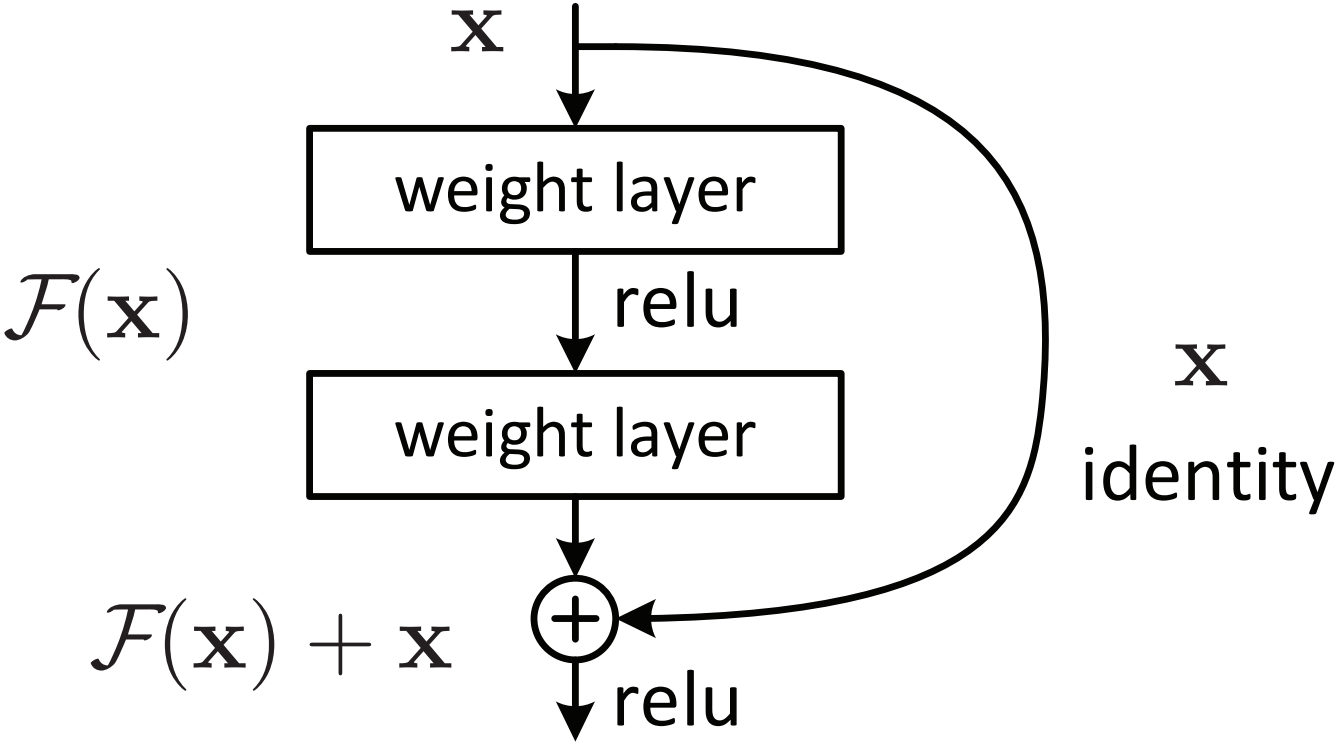}
\end{center}
\centering
\caption{Building block of ResNet \cite{2016Deep}}
\label{fig:Building-block-of-ResNet}
\end{figure}

\begin{figure}[t]
  \begin{center}
    \subfloat[If stride is 1]{\includegraphics[width=0.26\textwidth]{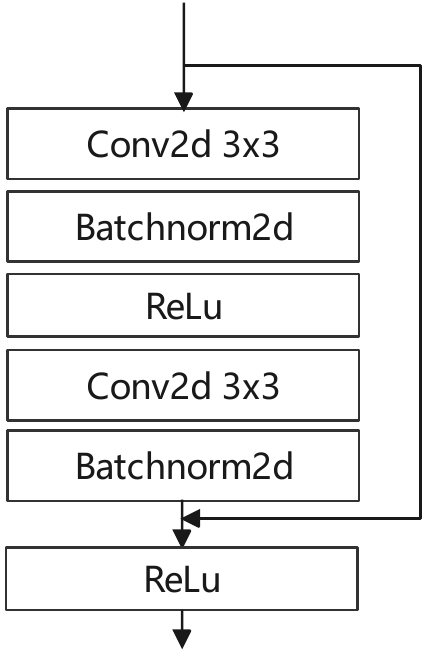}}
    \quad\quad\quad\quad
    \subfloat[If stride is 2 or input channels is not equal to output channels]{\includegraphics[width=0.42\textwidth]{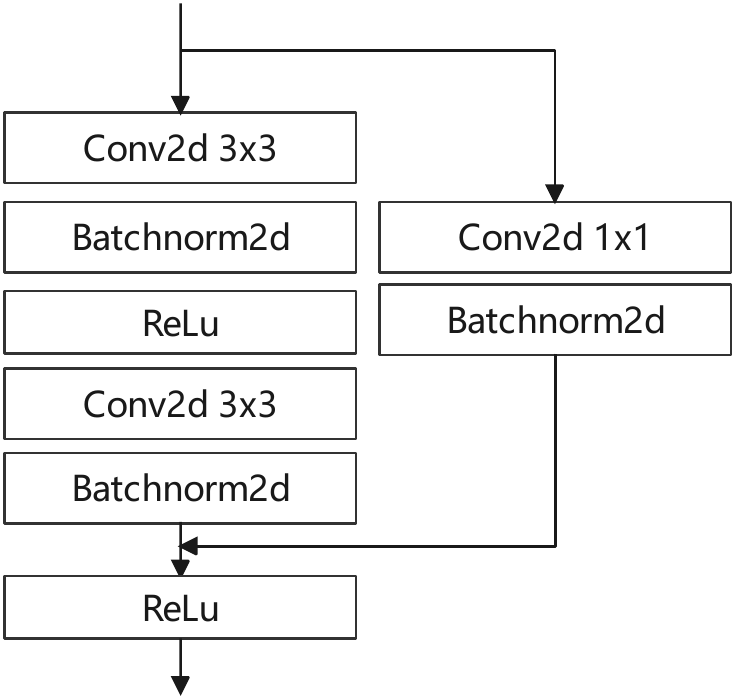}}
  \end{center}
  \caption{Structure of "Residual Block" in two conditions}
  \label{fig:structure-of-residualblock}
\end{figure}

ResNet-18 have 18 convolutional layers which is the smallest model of ResNet.
The paper~\cite{2016Deep} points out that the deep CNN is hard to optimize and
they introduce a network with residual short cut to remit this problem which
can deliver the top-level feature to the deep layers, and enable the training
of deep CNN. Fig.~\ref{fig:Building-block-of-ResNet} shows the building block of
ResNet. Table~\ref{The parameters configuration of ResNet-18} shows the parameters choose in our experiment.  Fig.~\ref{fig:structure-of-residualblock} shows the structure of ``Residual Block'' under two conditions.

\begin{table}[t]
\small
\centering
\begin{tabular}{l|l|l|l|l}
\hline
Module           & Input channels & Output channels & Stride & Features \\ \hline
Conv2d, Bn, ReLu & 1              & 64              & 1      & -        \\ 
ResidualBlock    & 64             & 64              & 1      & -        \\ 
ResidualBlock    & 64             & 64              & 1      & -        \\ 
ResidualBlock    & 64             & 128             & 2      & -        \\ 
ResidualBlock    & 128            & 128             & 1      & -        \\ 
ResidualBlock    & 128            & 256             & 2      & -        \\ 
ResidualBlock    & 256            & 256             & 1      & -        \\ 
ResidualBlock    & 256            & 512             & 2      & -        \\ 
ResidualBlock    & 512            & 512             & 1      & -        \\ 
Linear           & -              & -               & -      & 3        \\ \hline
\end{tabular}
\caption{The parameter’s configuration of ResNet-18}
\label{The parameters configuration of ResNet-18}
\end{table}

\subsubsection{{{MobileNetV1}}} 
\begin{figure}[t]
\begin{center}
\includegraphics[height=1.7cm]{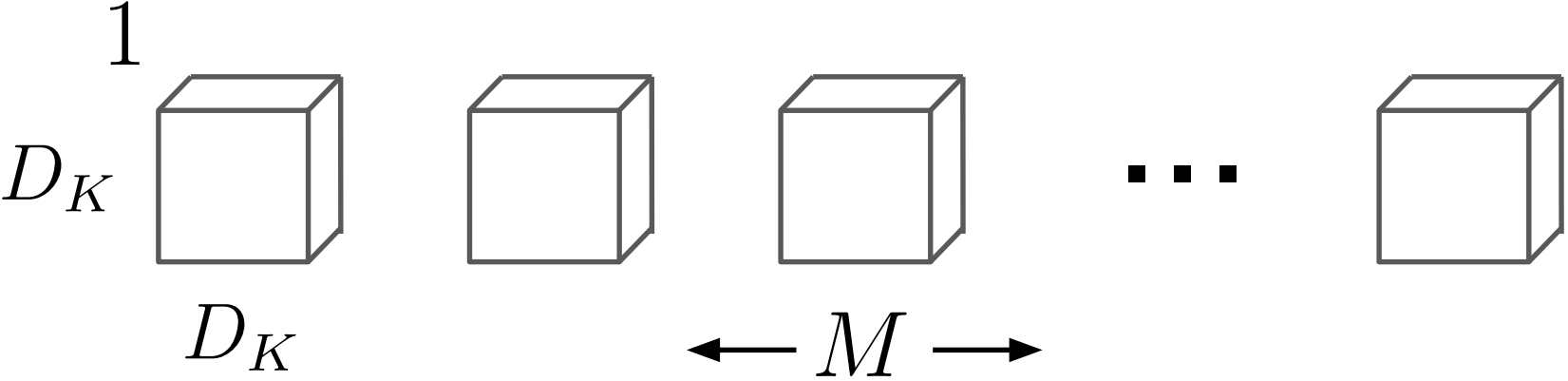}
\end{center}
\centering
\caption{Depthwise Convolutional Filters \cite{2017MobileNets}}
\label{fig:Depthwise-Convolutional-Filters}
\end{figure}

MobileNetV1 use depthwise separable convolution to reduce parameters amount and
increase inference speed which can enable the deployment of image recognition
algorithm on the mobile device. The depthwise convolution is a special case of
group convolution, the groups' number is the same as the input channels.
Fig~\ref{fig:Depthwise-Convolutional-Filters} shows the architecture of depthwise
convolutional filters in MobileNetV1. The paper~\cite{2017MobileNets} points
out that the computational cost of standard convolution can be denoted as:

\begin{equation}
\label{mobile1}
D_K \cdot D_K \cdot M\cdot N\cdot D_F\cdot D_F
\end{equation}
The computational cost of depthwise convolution can be denoted as:
\begin{equation}
\label{mobile2}
D_K \cdot D_K \cdot M\cdot D_F\cdot D_F
\end{equation}

In Formula \ref{mobile1} and \ref{mobile2},  denote the size of kernel, denote the size of the feature map, denote the input channels and denote the output channels. These formulae show that the computational complexity of depthwise convolution is lower than standard convolution.

\begin{figure}[t]
  \begin{center}
    \subfloat[BasicConv2d]{\includegraphics[width=0.26\textwidth]{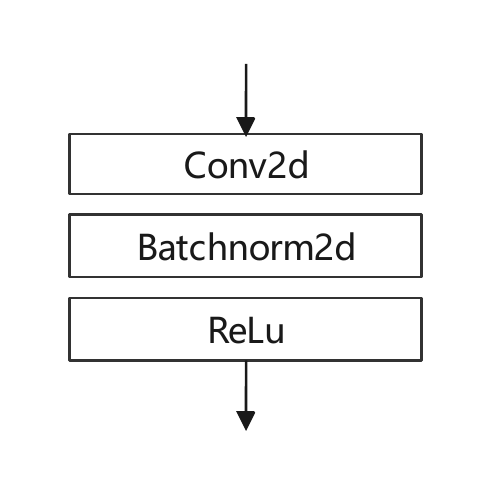}}
    \quad\quad\quad\quad
    \subfloat[DepthwiseConv2d]{\includegraphics[width=0.26\textwidth]{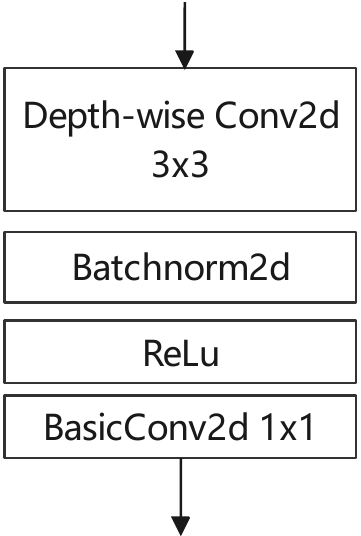}}
  \end{center}
  \caption{The structure of BasicConv2d and DepthwiseConv2d}
  \label{fig:BasicConv2d-DepthwiseConv2d}
\end{figure}

In our implementation, we use BasicConv2d and DepthwiseConv2d as basic modules of mobilenet.  Fig.~\ref{fig:BasicConv2d-DepthwiseConv2d} shows the structure of BasicConv2d and DepthwiseConv2d, and the parameter's configuration is given in 
Table~\ref{The parameters configuration of MobileNet}.

\begin{table}[t]
\small
\centering
\begin{tabular}{l|l|l|l|l|l}
\hline
Module          & Input Channels & Output Channels & Stride & Features & Repeat \\ \hline
BasicConv2d     & 1              & 32              & 1      & -        & 1      \\ 
DepthwiseConv2d & 32             & 64              & 1      & -        & 1      \\ 
DepthwiseConv2d & 64             & 128             & 2      & -        & 1      \\ 
DepthwiseConv2d & 128            & 128             & 1      & -        & 1      \\ 
DepthwiseConv2d & 128            & 256             & 1      & -        & 1      \\ 
DepthwiseConv2d & 256            & 256             & 1      & -        & 1      \\ 
DepthwiseConv2d & 256            & 512             & 2      & -        & 1      \\ 
DepthwiseConv2d & 512            & 512             & 1      & -        & 5      \\ 
DepthwiseConv2d & 512            & 1024            & 2      & -        & 1      \\ 
DepthwiseConv2d & 1024           & 1024            & 1      & -        & 1      \\ 
Avgpool2d 4x4   & -              & -               & -      & -        & 1      \\ 
Linear          & -              & -               & -      & 3        & 1      \\ \hline
\end{tabular}
\caption{The parameter’s configuration of MobileNet}
\label{The parameters configuration of MobileNet}
\end{table}

\section{Results \& Discussion}
\label{sec:Results}
\subsection{Subjective Evaluation}
Table~\ref{Results of subjective evaluation} shows the results of the subjective evaluation. We compute several objective metrics~\cite{dev2015multi} -- accuracy, Gini index, Adjusted F-score, and Area Under Curve for benchmarking purposes. The following is our explanation of each metric used in the experiment:

\begin{table}[t]
\footnotesize
\centering
\begin{tabular}
{p{2.4cm}||p{0.6cm}|p{3.1cm}|p{2.6cm}|p{2.5cm}}
 \hline
 Method & ACC & Gini Index & AUC & AGF\\
 \hline
CNN  &37\% &0:-0.11,1:0.38,2:0.012 &0:0.44,1:0.69,2:0.5 &0:0.40,1:0.63,2:0.50 \\
HOG+SVM  &58\% &0:0.2,1:0.5, 2:0.19 &0:0.6,1:0.75,2:0.59 &0:0.66,1:0.71,2:0.52 \\
HOG+CNN  &42\% &0:0.33,1:0.083,2:0.05 &0:0.66,1:0.54,2:0.43 &0:0.62,1:0.42,2:0.3 \\
HOG+SIFT+SVM  &32\% &0:-0.02,1:-0.05,2:-0.002 &0:0.49,1:0.47,2:0.49 &0:0.44,1:0.45,2:0.46 \\
SURF+SVM  &35\% &0:0.0005,1:0.083,2:0.012 &0:0.5,1:0.54,2:0.51&0:0.47,1:0.52,2:0.47 \\
DenseNet-121  &78\% &0:0.6638,1:0.34,2:0.46 &0:0.83,1:0.67,2:0.73&0:0.81,1:0.69,2:0.73 \\
ResNet-18  &80\% &0:0.78,1:0.37,2:0.499 &0:0.89, 1:0.69,2:0.75&0:0.89,1:0.68,2:0.77 \\
MobileNetV1 &66\% &0:0.237,1:0.088,2:0.19 &0:0.62,1:0.54,2:0.597 &0:0.51,1:0.512:0.65 \\
 \hline
\end{tabular}
\caption{Results of subjective evaluation}
\label{Results of subjective evaluation}
\end{table}

\textbf{1. Accuracy (ACC)}: total number of images in test data be $\tau$ and the total number of correct predictions be $\Omega$, accuracy can be expressed as:
\begin{equation}
ACC = \frac{\Omega}{\tau}
\end{equation}
\textbf{2. Gini Index}: the Gini Index is determined by deducting the sum of squared of probabilities of each class from one, mathematically, Gini Index can be expressed as:
\begin{equation}
 GI = 1 -  \sum_{i=1}^{n} (P_{i})^2
\end{equation}
where $P_i$ denotes the probability of an element being classified for a distinct class.

\textbf{3. Adjusted F-Score (AGF)}:  computed by taking the geometric mean of $F_2$ and $inv F_{0.5}$ where:
\begin{equation}
 F_2 = \frac{5}{4} \times \frac{sensitivity\times precision}{ sensitivity + precision} 
\end{equation}
\begin{equation}
 inv F_{0.5} = \frac{5}{4} \times \frac{sensitivity \times precision}{0.5^2 \times sensitivity + precision}
\end{equation}
\begin{equation}
 AGF = \sqrt{F_2 \times inv F_{0.5}}
\end{equation}

In this experiment, we use sensitivity and precision to evaluate the performance of the model. The following formulae are the ones for calculating sensitivity and precision. 
Table~\ref{Meaning of some terminology's abbreviation} shows the meaning of some terminology's abbreviation in Formula. 
\begin{equation}
 sensitivity  = \frac{TP}{TP + TN}
\end{equation}
\begin{equation}
 precision =  \frac{TP}{TP + FP}
\end{equation}

\begin{table}[t]
\centering
\begin{tabular}{c|l}
\hline
Abbreviation & \multicolumn{1}{c}{Meaning} \\ \hline
TP           & True Positive                \\ 
TN           & True Negative                \\ 
FP           & False Positive               \\ 
FN           & False Negative               \\ \hline
\end{tabular}
\caption{Meaning of some terminology's abbreviation}
\label{Meaning of some terminology's abbreviation}
\end{table}

\textbf{4. Area Under Curve  (AUC)}: We also compute the area under ROC (receiver operating characteristic) curve for the various classes. The AUC assists us in visualizing the trade-off between true positive rate (TPR) and false positive rate (FPR). The TPR and FPR are computed for various thresholds and plotted in a single plot to compute the AUC curve. 

Next we explain in detail about the dimensionality reduction techniques that have been used to understand the reason behind the low accuracy.

\subsection{Principal Component Analysis}

Amongst the several feature selection techniques~\cite{dev2016rough}, the Principal Component Analysis (PCA) is widely used. The principal components in PCA capture the most variation in a dataset. PCA deals with the curse of dimensionality by capturing the essence of data into a few principal components. Now we define the PCA Algorithm:

 \textbf {1. }Given ($x_1,x_2$, \ldots ...., $x_m$) in $R^n$ we try to represent them in $R^l$ where $l<n$.

Each $x_i$ represents a vector that we have created from our training images. In our case, $n$ is 30\,000 and $l$ is 2 that is we bring down a vector from $R^{30000}$ to $R^2$.

 \textbf {2. }We aim to find what we call as an encoder function and a decoder function to take our $x_i$ from $n$ to $l$ and from $l$ to $n$ respectively.
Now define:
\[
 \left\{
\begin{aligned}
f(x) = c, \text{ where f is the encoder function} \\
g(c) = x, \text{ where g is the decoder function}
\end{aligned}
\right.
\]

One of the decoding function is $g(c) = Dc$, where $D$ is a matrix with $l$ mutually orthogonal columns to minimize distance between $x$ and the reconstruction $r(x) = g(f(x)) = DD^Tx$. The optimal $c^{*}$ and $D^{*}$ are given as:
\begin{equation}
c^{*} = \argmin_c = \Vert x - g(c)\Vert _2
\end{equation}

\begin{equation}
D^* = \argmin_D[\sum_{i,j}(x_j^i - r(x^i)_j)^2)]^{1/2}
\end{equation}

After solving the above equations we observe that $D$ is a matrix with top $l$ eigenvectors of design matrix X $\epsilon$ $R^{mxn}$.

\begin{figure}[htb]
\begin{center}
\includegraphics[height=7.5cm]{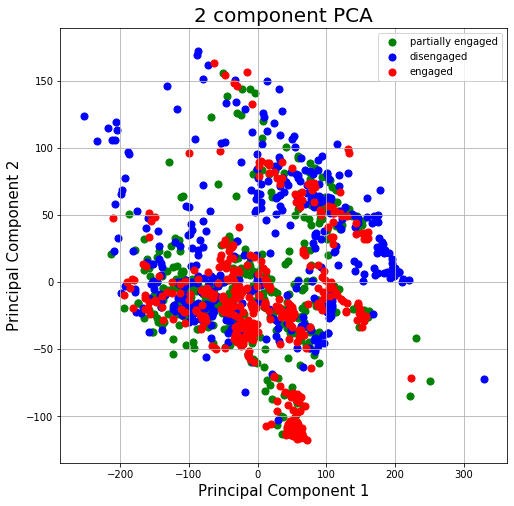}
\end{center}
\centering
\caption{We show the distribution of samples in the lower-dimensional subspace generated by principal component analysis. We observe that the different labels are not well-separated indicating the difficulty of engagement detection.}
\label{fig:pca}
\end{figure}

\subsection{T-{{SNE}}}

t-SNE is a non-linear dimensionality reduction algorithm used for exploring high-dimensional data. It maps multi-dimensional data to two or more dimensions suitable for human observation. The details of the t-SNE algorithm are as follows.

\textbf{1. }Given a set of $N$ high dimensional objects $x_1,x_2, \ldots ..x_N$ t-SNE first computes the probability $P_{i \vert j}$ that are proportional to the similarity of objects $x_i$ and $x_j$ as follows. For $i\neq j$, define:
\begin{equation}
    p_{i\mid j}= \frac{exp(-\left | \left | x_i - x_j \right | \right |^{2} / 2\sigma _i^{2})}{\sum _{k\neq i}exp(-\left | \left | x_i - x_j \right | \right |^{2} / 2\sigma _i^{2})}
\end{equation}
And set $P_{i \vert i} = 0$. Note that $\sum _{j}p_{j\mid i} = 1$ for all $i$. Now define:
\begin{equation}
p_{ij} = \frac{p_{j\mid i} + p_{i\mid j}}{2N}
\end{equation}

The similarity of datapoint $x_j$ to datapoint $x_i$ is the conditional probability, $P_{j \vert i}$, that $x_i$ would pick $x_j$ as its neighbor if neighbors were picked in proportion to their probability density under a Gaussian centered at $x_i$.

\textbf{2. }t-SNE aims to learn a {\bf d}-dimensional map $y_1,y_2, \ldots ..y_N$ ($y_i\epsilon R^d$) that reflects the similarities $P_{i,j}$ as well as possible. To this end, it measures similarities $Q_{i,j}$ between two points in the map $\mathbf{y}_{\mathbf{i}}$ and $\mathbf{y}_{\mathbf{j}}$ using a very similar approach. Specifically, for $i\neq j$, define $Q_{i,j}$ as:
\begin{equation}
Q_{ij} = \frac{(1+\left | \left | y_i - y_j \right | \right |^{2})^{-1}}{\sum _k\sum _{\xi \neq k} (1+\left | \left | y_i - y_\xi  \right | \right |^{2})^{-1}}
\end{equation}
And set $Q_{i,i} = 0$.

The locations of the points $\mathbf{y}_{\mathbf{i}}$ in the map are determined by minimizing the (non-symmetric) KL Divergence of the distribution {\bf P} from the distribution {\bf Q}, that is:
\begin{equation}
KL\left ( P\vert\vert Q \right )= \sum {\mbox{}_{i\neq j}} P_{i,j} log\left (\frac{P_{i,j}}{Q_{i,j}} \right )
\end{equation}

The minimization of the Kullback--Leibler divergence with respect to the points $\mathbf{y}_{\mathbf{i}}$ is performed using gradient descent. The result of this optimization is a map that reflects the similarities between the high-dimensional inputs.

\begin{figure}[htb]
\begin{center}
\includegraphics[height=7.5cm]{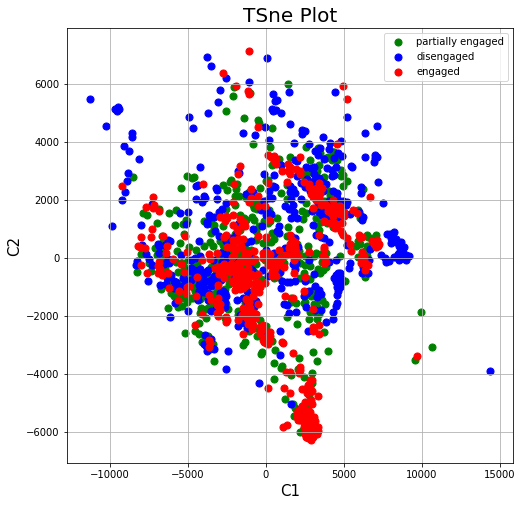}
\end{center}
\centering
\caption{We show the distribution of samples in the lower-dimensional subspace generated by t-distributed stochastic neighbor embedding. Similarly, we also observe that the different labels are not well-separated indicating the difficulty of engagement detection.}
\label{fig:tsne}
\end{figure}

\subsection{Reasons of low performance}

In Fig.~\ref{fig:pca} and Fig.~\ref{fig:tsne}, we can see that the distribution followed by partially engaged and engaged is very similar, so our techniques cannot differentiate between an image that is partially engaged and engaged. This tells us that there are no hard boundaries that we can define which clearly separates the two classes from each other but we can see that the distribution of disengaged and engaged is very different so it is easy to classify between them. The accuracy that we get is not very high and one main reason for this is the distribution of data points. From the plots, we can see that the distribution of engaged and partially engaged are nearly identical hence resulting in low accuracy.

\subsection{Discussion}

\begin{figure}[htb]
\begin{center}
\includegraphics[height=8cm]{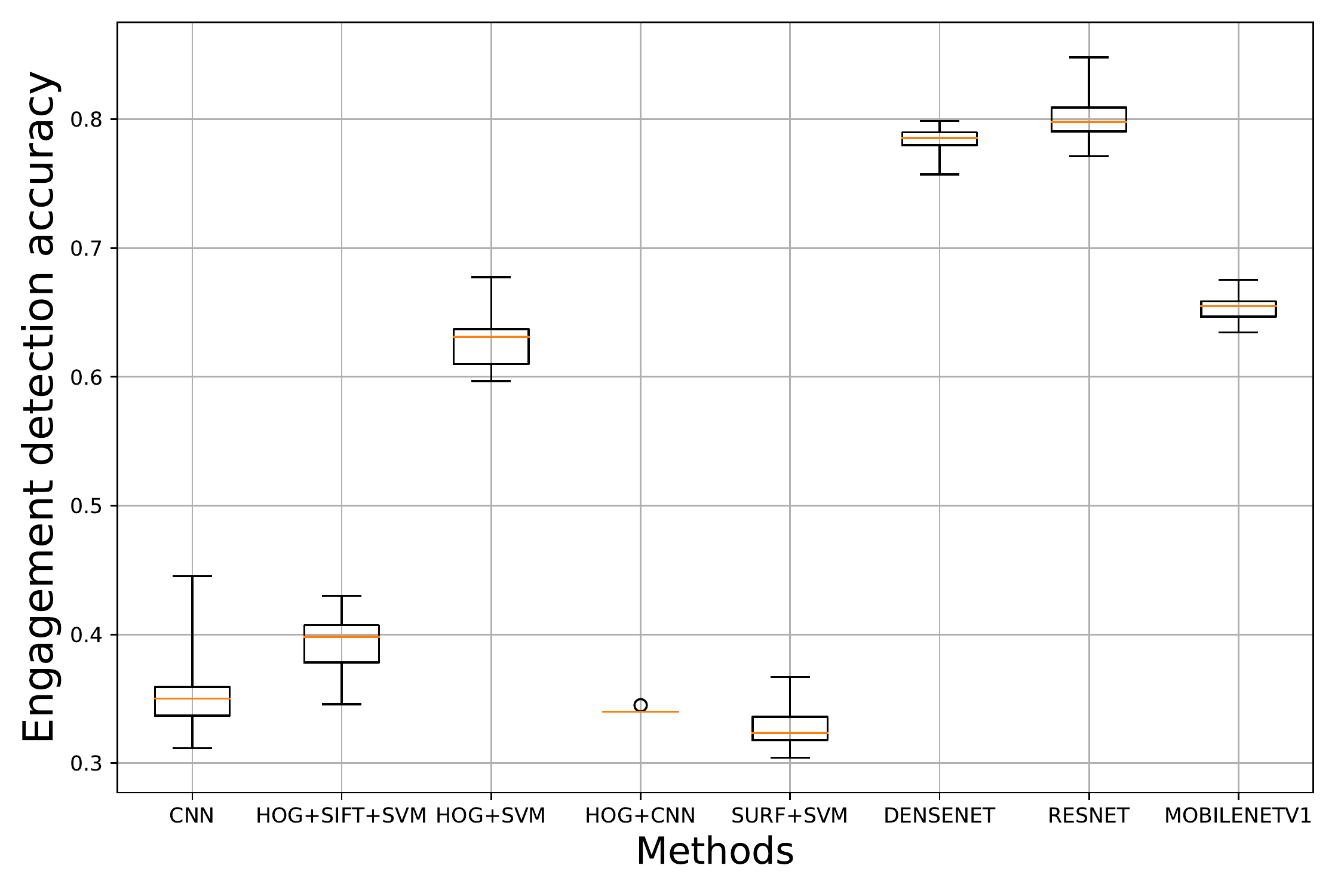}
\end{center}
\centering
\caption{We show the distribution of engagement detection accuracy for the different benchmarking methods. We perform $10$ random experiments for all the different benchmarking methods.}
\label{fig:boxplots}
\end{figure}

In  Fig.~\ref{fig:boxplots}, in terms of the non-deep learning method, we can see the range of accuracies obtained by different methods. In the case of HOG$+$CNN the accuracy for each experiment is nearly the same and is equal to 34\%. We get the highest accuracy for HOG$+$SVM technique of 63\%. Other techniques are more or less are working in the same manner but HOG$+$SVM performs much better. In terms of the deep learning method, both ResNet-18 and DenseNet-121 have good accuracy performance with around 80\% and 78\% respectively. The performance of MobileNetV1 is 66\% which is a little better than HOG$+$SVM.

\section{Conclusions and future work}

\label{sec:5}
In this paper, we explored different techniques of extracting features from images and train our models on those features. We used different feature descriptors and keypoint detectors being used to generate training vectors. We see that PCA and t-SNE tell us about the representation of our data which helps us to analyze why our techniques are working in this manner. Our results indicate that the deep-learning methods have better performance, with ResNet-18 possessing the best accuracy performance among three deep-learning models. Amongst the non-deep-learning methods, HOG$+$SVM has the best performance. We recommend online education platforms to reasonably select deep- or non-deep-learning methods according to their own conditions and needs. The limitations of our research is that our techniques cannot differentiate between a human that is partially engaged and fully engaged because there are no clear and hard boundaries that we can be defined which clearly separates the two classes.  

For future work, we intend to further study the deep-learning methods on engagement detection and work on various other datasets to further benchmark our proposed model. Especially, we will focus on combining advanced deep-learning models' block such as visual transformer and RepVGG. We also plan to embed the proposed algorithm model into the online education platform to observe its practical role.

\section*{Acknowledgments}
This research was conducted with the financial support of SFI Research Centres Programme under Grant 13/RC/2106\_P2 at the ADAPT SFI Research Centre at University College Dublin. ADAPT, the SFI Research Centre for AI-Driven Digital Content Technology, is funded by Science Foundation Ireland through the SFI Research Centres Programme.

\balance

\end{document}